# Pyramidal RoR for Image Classification

Ke Zhang[1], Liru Guo[1], Ce Gao[1], Zhenbing Zhao[1]

*1. Department of Electronic and Communication Engineering, North China Electric Power University, Baoding, China*
*Eail：zhangke41616@126.com*

**Abstract:**

The Residual Networks of Residual Networks (RoR) exhibits excellent performance in the image classification task, but sharply increasing the number of feature map channels makes the characteristic information transmission incoherent, which losses a certain of information related to classification prediction, limiting the classification performance. In this paper, a Pyramidal RoR network model is proposed by analysing the performance characteristics of RoR and combining with the PyramidNet. Firstly, based on RoR, the Pyramidal RoR network model with channels gradually increasing is designed. Secondly, we analysed the effect of different residual block structures on performance, and chosen the residual block structure which best favoured the classification performance. Finally, we add an important principle to further optimize Pyramidal RoR networks, drop-path is used to avoid over-fitting and save training time. In this paper, image classification experiments were performed on CIFAR-10/100 and SVHN datasets, and we achieved the current lowest classification error rates were 2.96%, 16.40% and 1.59%, respectively. Experiments show that the Pyramidal RoR network optimization method can improve the network performance for different data sets and effectively suppress the gradient disappearance problem in DCNN training.

**Keywords:** image classification; Residual Networks of Residual Networks; PyramidNet; Pyramidal RoR

## 1. Introduction

In the past five years, deep learning [1, 2] has made gratifying achievements in various computer vision tasks. With the rapid development of deep learning and Convolutional Neural Networks(CNNs), image classification has bidden farewell to coarse feature problems of manual extraction, and turned it into a new process. Especially, after AlexNet [13] won the champion ship of the 2012 Large Scale Visual Recognition Challenge(ILSVRC) [4], CNNs become deeper and continue to achieve better and better performance on different tasks of computer vision tasks.

Since AlexNet acquired a celebrated victory at the ImageNet competition in 2012, convolution neural network has witnessed a great development in the past few years. But the deep convolutional neural networks (DCNNs) suffer from gradient vanish, degradation and other problems, which have all been restricting its development and application.

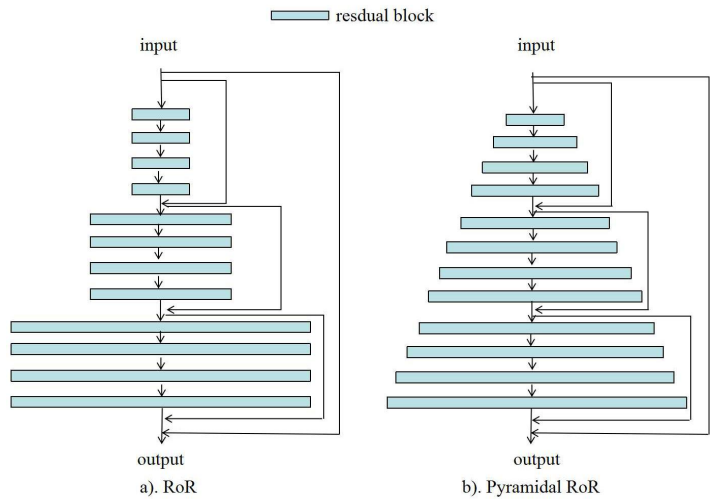

Figure. 1 Pyramidal RoR architecture.

To overcome degradation problems, a residual learning framework named Residual Networks (ResNets) were developed [8] to ease network, which achieved excellent results on the ImageNet test set. Since then, current state-of-the-art image classification systems are predominantly variants of ResNets. Residual networks of Residual networks (RoR) [21] adds level-wise shortcut connections upon original residual networks to promote the learning capability of residual networks. RoR exhibits excellent performance in the image classification task. However, sharply increasing the number of feature map channels in RoR makes the characteristic information transmission in the network incoherent, which losses a certain of information related to classification prediction, and limits the classification performance.

To effectively solve the above problem, this paper proposes an RoR network optimization method as shown in Figure 1. In this paper, by analysing the performance characteristics of RoR and combining with the PyramidNet, a Pyramidal RoR network model framework is proposed. Firstly, based on RoR, the Pyramidal RoR network model with channels gradually increasing is designed. Secondly, we analysed the effect of different residual block structure on performance, and chosen the residual block structures which best favoured the classification performance. Finally, we add an important principle to further optimize Pyramidal RoR networks. Drop-path is used to avoid over-fitting and save training time.

## 2. Background

Since AlexNet acquired a celebrated victory at the ImageNet competition in 2012, an increasing number of deeper and deeper Convolutional Neural Networks emerged, such as the 19-layer VGG [7] and 22-layer GoogleNet [7]. However, very deep CNNs also introduce new challenges: degradation problems, vanishing gradients in backward propagation and overfitting [5].

To overcome the degradation problem, a residual learning framework known as residual networks (ResNets) [8] was presented at the ILSVRC 2015 & COCO 2015 competitions to ease the training of networks, and achieved excellent results in combination with the ImageNet test set. Since then, a series of optimized models based on ResNets has emerged, which became part of the Residual-Networks Family. In the Pre-ResNets, He et al. [12] created a direct path for propagating information through the entire network, which made training easier and improved generalization. Inspired by Pre-ResNets, Shen et al. [15] proposed weighted residuals for very deep networks (WResNet), which removed the ReLU from highway and used weighted residual functions to create a direct-path. This method is also capable of 1000+ layers residual networks training and achieves good accuracy. To further reduce vanishing gradients. To prevent overfitting. Huang and Sun et al. [13] proposed a drop-path method, the stochastic depth residual networks (SD), which randomly drops a subset of layers and bypassed them with identity mapping for every mini-batch. To tackle the problem of diminishing feature reuse, wide residual networks (WRNs) [14] were introduced by decreasing depth and increasing width of residual networks.

Even though non-saturated rectified linear unit (ReLU) has interesting properties, such as sparsity and non-contracting first-order derivative, its non-differentiability at the origin and zero gradient for negative arguments can hurt back-propagation [24]. Moreover, its non-negativity induces bias shift causing oscillations and impeded learning. Since the advent of the well-known ReLU, many have tried to further improve the performance of the networks with more elaborate functions. "Exponential linear unit" (ELU) [24], defined as identity for positive arguments and *exp(x)-1* for negative ones, deals with both increased variance and bias shift problems. Parametric ELU (PELU) [25], an adaptive activation function, defines parameters controlling different aspects of the function and proposes learning them with gradient descent during training.

Residual networks of Residual networks (RoR) [21] adds level-wise shortcut connections upon original residual networks to promote the learning capability of residual networks, that once achieved state-of-the-art results on CIFAR-10 and CIFAR-100 [19]. Instead of sharply increasing the feature map dimension at units that perform downsampling, PyramidNet [32] gradually increase the feature map dimension at all units and has superior generalization ability. DenseNet [22] uses *a densely connected path* to concatenate the input features with the output features, enabling each micro-block to receive raw information from all previous micro-blocks. Wang et al [36] proposed "Residual Attention Network", a convolutional neural network using attention mechanism which can incorporate with state-of-art feed forward network architecture in an end-to-end training fashion. To enjoy the benefits from both path topologies of ResNets and DenseNet, Dual Path Network [35] shares common features while maintaining the flexibility to explore new features through dual path architectures.

## 3. Methodology

In this paper, firstly, the basic structure of RoR is given. Secondly, by analysing the characteristics of RoR network and PyramidalNet, a Pyramidal RoR is designed. Finally, two different residual blocks are analysed and drop-path is used to avoid over-fitting for optimization.

### 3.1 RoR network architecture

To further improve the learning ability of ResNets, Zhang et al. [21] hypothesized that if the residual mapping is easier to learn, the residual mapping of the residual mapping should be easier to learn. Based on ResNets, RoR networks is set up. The structure shown in Figure 2. RoR adds multi-level shortcuts based on ResNets. The shortcut on the left is a root-level shortcut, and the remaining shortcuts are made up of three orange shortcuts, which are middle-level shortcuts. The blue shortcuts are final-level shortcuts. Therefore, high-level residual blocks can transfer information to the underlying residual blocks, which plays an important role in suppressing the gradient vanish. Experiments showed that RoR has obtained the best classification results on CIFAR-10, CIFAR-100 and SVHN. Figure 2 shows the RoR architecture. The optimal model is 3-level RoR in [21]. Therefore, we adopted 3-level RoR (RoR-3) as our basic architecture in experiments. RoR-3 has three-level residual blocks, which are root level, middle level and final level residual blocks as shown in Figure 2.

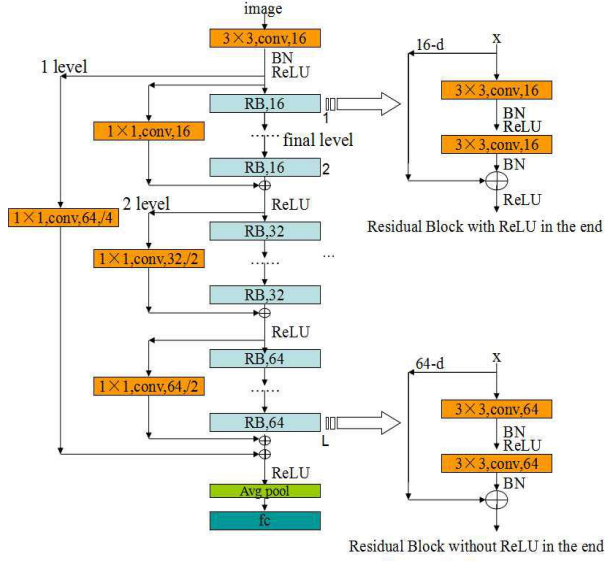

Figure 2 Residual Networks of Residual Networks

### 3.2 Pyramidal RoR

The same as the most DCNN architectures [6, 8, 12, 13, 14, 21, 24, 25, 32], RoR follows a principle that the number of channels is sharply increased at downsampling locations. In the case of RoR on the CIFAR data set [12], the number of feature map dimensions $Dk$ of the $k$-th residual blocks that belongs to the $n$-th group can be described as follows: the number of channels and the feature maps in each group of residual blocks are the same. At the beginning of the next group of residual blocks, size of feature maps is halved and the number of feature map channels is doubled, as in (1). Where the number of residual blocks groups is $n \subseteq \{1, 2, 3\}$ in the RoR of the CIFAR data set [12].

$$D_n = 16^n \qquad (1)$$

That can not only reduce dimensionality of feature maps, but also increases the diversity of the high-level attributes in high level layers by doubling the feature map channels. However, sharply increasing the number of feature map channels makes the characteristic information transmission in the network incoherent, which losses some relevant information related to classification prediction, and limits the classification performance. PyramidNet [29] concentrates on the feature map dimension by increasing it gradually instead of by increasing it sharply at each residual unit with downsampling. The utilization of gradually increasing the number of feature map channels to ensure the diversity of advanced attributes, while ensuring the continuity of information.

Inspired by PyramidNet, we propose a Pyramidal RoR network model, which contains multi-level shortcuts. So that different layers of information can be transmitted to each other, and increasing the feature dimension gradually makes information transmission more smoothly. we employ a method of increasing the feature map dimension as follows:

$$D_n = \begin{cases} 16, & \text{if } n = 1, \\ \lfloor D_{n-1} + \alpha/N \rfloor, & \text{if } 2 \leq n \leq N+1, \end{cases} \qquad (2)$$

In which $N$ denotes the total number of residual blocks, defined as $N = 3n$. RoR contains 3 group residual blocks on CIFAR dataset, and each group contains $n$ residual blocks. The dimension is increased by a step factor of $\alpha / N$, and the output dimension of the final block of the final group becomes $16 + \alpha$.

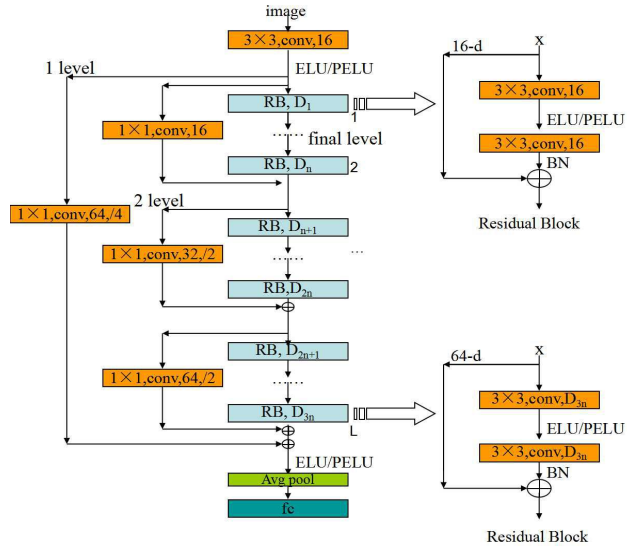

Figure.3 Pyramidal RoR architecture.

We construct Pyramidal RoR as shown in Figure 3. First, three groups final-level residual blocks are stacked with same numbers in Pyramidal RoR network. The convolution layer before the three sets of residual blocks is defined as Conv 1. The three groups are Conv 2, Conv 3 and Conv 4. The *k* th residual block in each group is defined as Conv 2_k. The channels of convolutional layers in 3 groups final-level residual blocks are gradually increased linearly. Input of Conv 2_1 is 16 channels and the output of Conv 2_1 is increased by $\lfloor \alpha / 3n \rfloor$. Which means, subsequently, channels in each residual block increased by $\lfloor \alpha / 3n \rfloor$, until channels increased to $D_{3n} = 16 + \alpha$ in the last one. The feature map size in three groups is 32×32, 16×16, 8×8, respectively. Downsampling locations are the first convolution layer in the 2ed and 3th groups. Pyramidal RoR includes *3n* final residual blocks, 3 middle-level residual blocks, and a root-level residual block, among which a middle-level residual block is composed of final residual blocks and a middle-level shortcut, the root-level residual block is composed of 3 middle-level residual blocks and a root-level shortcut. ReLU is followed by an addition. The projection shortcut is done by 1×1 convolutions.

## 4.Pyramidal RoR Optimization Method

In this section, we present an in-depth study of the architecture of our Pyramidal RoR, together with two types of residual units, and Stochastic Depth. The experiments we include here support the study and confirm that insights obtained from our network architecture can further improve the performance of existing RoR-based architectures.

### 4.1 Structure of Residual Blocks

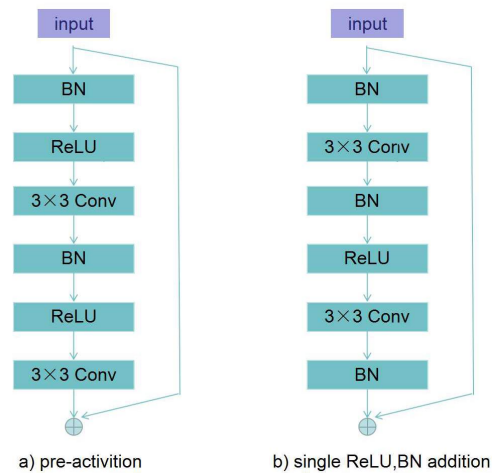

Figure.4 Architecture of two residual blocks

Pyramidal RoR is like ResNets, which is composed of residual blocks (convolutional filter, ReLU, BN layer stacking and shortcuts) as the basic structure of the accumulation. Therefore, the residual block is the core of Pyramidal RoR. Structure of residual blocks directly determines the image classification performance of the network. In this section, two different residual blocks are discussed, and most suitable residual block is determined for Pyramidal RoR classification performance.

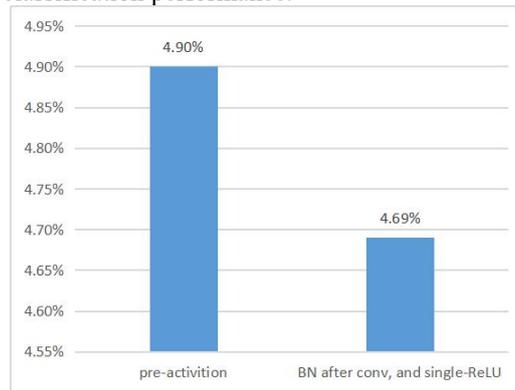

Figure.5 Performance comparison of two residual blocks.

As shown in Figure. 4, a) is the residual block structure used in Pre-ResNet [12]. The activation function is removed from the trunk network which is between two adjoining residual blocks, resulting in an identity mapping. The identity mapping is more conducive to information transmission compared to the original residual block; a) uses the BN-ReLU-Conv order. ReLU provides the non-linearity for the network as an activation function. However, it filters out negative values, so the negative value is filtered after the convolution layer. The input of the residual block is positive and lossy. Pre-ResNet [12] uses the BN-ReLU-Conv order, and the convolution layers play an important role in providing negative values. The residual block b) [34] is the structure of BN-Conv-BN-ReLU-Conv-BN. The study shows that too many ReLU can impair network performance. BN [18] plays a role in regulating the activation function and accelerating convergence in the network. [15] increased the BN structure in the last of residual block and played a good effect, so that [34] used BN-Conv-BN-ReLU-Conv-BN residual block structure.

Two different residual blocks are applied to the Pyramidal RoR model for image classification experiments on CIFAR-10. 110-layer Pyramidal RoR networks with two different residual blocks, are trained on CIFAR-10 training dataset after 500 Epoch, which get effective convergence with a time of 12.2h. It can be seen from Figure 5, The error rate of classification on the test set using b) residual block is 4.69%, which is significantly lower than that using a) residual block. Therefore, in the subsequent test, we use b) residual block as basic blocks.

**4.2 Stochastic Depth**

Pyramidal RoR networks widens the network and adds more training parameters while adding additional levels shortcuts, which can lead to more serious overfitting problems. The most frequently applied algorithms to overcome overfitting are dropout [30, 17] and drop-path [13], which modify interactions between sequential network layers for discourage co-adaptaion. Dropout is less effective when used in convolutional layers and ResNets [14]. We give up dropout in Pyramidal RoR. Drop-paths prevent co-adaptation of parallel paths by randomly dropping the path. He et al. [12] proved that the network cannot converge to a good solution by dropping an identity mapping path randomly, because dropping an identity mapping path greatly influences training. However, Huang et al. [13] proposed a Stochastic Depth drop-path method which only dropped the residual mapping path randomly. Their experiments showed that the method reduced the test errors significantly.

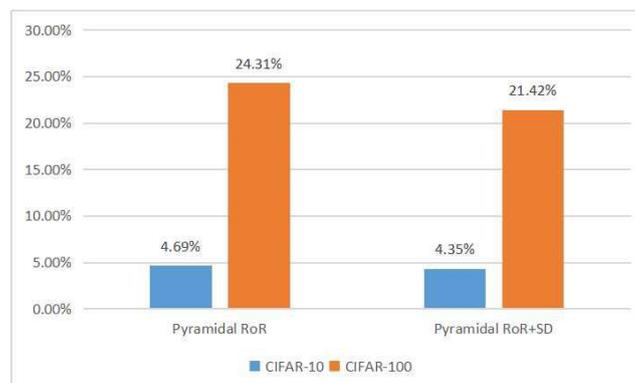

Figure. 6 Performance comparison of Stochastic Depth

Table 1 Training time comparison on CIFAR-10/100

|  | Pyramidal RoR | Pyramidal RoR+SD |
|---|---|---|
| CIFAR-10/100 | 12.2h | 6.9h |

Therefore, we used the Stochastic Depth (SD) algorithm, which is commonly used in residual networks, to alleviate the overfitting problem. We trained our Pyramidal RoR networks by randomly dropping entire residual blocks during training and bypassing their transformations through shortcuts, without performing forward-backward computation or gradient updates. Let $p_l$ mean the probability of the unblocked residual mapping branch of the $l$ th residual block. $L$ is the number of residual blocks, and (4) shows that $p_l$ decreases linearly with the residual block position. $p_L$ indicates that the last residual block is probably unblocked. SD can effectively prevent overfitting problems and reduce training time.

$$p_l = 1 - \frac{l}{L}(1 - p_L) \qquad (4)$$

The same layers and output dimensions of the two networks are training on CIFAR-10 and CIFAR-100, and the test results show in Figure. 6. Training time comparison shows in Table 1. Each type of object in CIFAR-100 contains 600 images and relatively few. Therefore, the network training with same order of magnitude parameters, over-fitting problem is more serious. As can be seen from Figure 6, training with SD can further enhance the Pyramidal RoR classification performance, and effectively control the over-fitting problem on CIFAR-10 and CIFAR-100. In the training phase, after 500 Epoch, the network is effectively convergent, with 12.2h without SD. The network with SD consumed 6.9 hours, and save 40% of time.

## 5. Experiment

To analyze the characteristics of Pyramidal RoR, as well as verify the effectiveness of the optimization scheme, massive experiments were planned. The implementation and results follow.

### 5.1 Implementation

In this paper, we used Pyramidal RoR for image classification, in two image datasets, CIFAR-10 and CIFAR-100 and compared with current excellent methods [21, 34]. CIFAR-10 contains 10 classes of objects, and CIFAR-100 contains 100 classes of objects. 50000 images of each dataset were used for network training. The remaining 10000 of that are used for testing classification performance. Our implementations were based on Torch 7 with a Titan X. We initialized the weights as in [10]. In both CIFAR-10 and CIFAR-100 experiments, we used SGD with a mini-batch size of 128 for 500 epochs. The learning rate started from 0.1, turned into 0.01 after epoch 250 and to 0.001 after epoch 375. In SVHN experiments, we used SGD with a mini-batch size of 32 for 50 epochs. The learning rate started from 0.1, turned into 0.01 after epoch 30 and to 0.001 after epoch 35. For the SD drop-path method, we set $p_l$ with the linear decay rule of $p_0 = 1$ and $p_L=0.5$. Other architectures and parameters were the same as RoR's. As for the data size being limited in this paper, the CIFAR experiments adopted two kinds of data expansion techniques: random sampling and horizontal flipping.

Table 2 Parameters in each layer of Pyramidal RoR

| Group | Output size | Building Block |
|---|---|---|
| Conv 1 | 32×32 | $[3\times3, 16]$ |
| Conv 2 | 32×32 | $\begin{bmatrix} 3\times3, \lfloor 16+\alpha(k-1)/N \rfloor \\ 3\times3, \lfloor 16+\alpha(k-1)/N \rfloor \end{bmatrix} \times N_2$ |
| Conv 3 | 16×16 | $\begin{bmatrix} 3\times3, \lfloor 16+\alpha(k-1)/N \rfloor \\ 3\times3, \lfloor 16+\alpha(k-1)/N \rfloor \end{bmatrix} \times N_3$ |
| Conv 4 | 8×8 | $\begin{bmatrix} 3\times3, \lfloor 16+\alpha(k-1)/N \rfloor \\ 3\times3, \lfloor 16+\alpha(k-1)/N \rfloor \end{bmatrix} \times N_4$ |
| acg pool | 1×1 | $[8\times8, 16+\alpha]$ |

Parameters in each layer and other details of Pyramidal RoR are shown in Table 2. Where $\alpha$ is the output network width factor, $N_k$ is the number of residual blocks contained in each group, and the first convolutional layer of Conv 2 and Conv 3 (Conv 2_1 and Conv 3_1) are reduced by convolution of step 2 sampling. The first and second level shortcuts are of type B (1 × 1 convolution) [8], [21], and the shortcut of final residual blocks are of type A [8], [21].

**5.2 Effect of Pyramidal RoR**

In this work, the TOP-1 error rate is used to evaluate the proposed network architecture. Change the output dimension, the experimental results are as follows:

The 110-layer PyramidNet, RoR and Pyramidal RoR ($\alpha$ = 48) were compared with 64 channels output, and the classification error rates on CIFAR-10 are shown in Figure. 7. In Figure. 7, the 110-layer RoR without SD resulted in a competitive 5.08% error on the test set, and PyramidNet resulted in 5.15%. The 110-layer Pyramidal RoR without SD had a 4.90% error on the test set and outperformed the 110-layer RoR without SD and PyramidNet on CIFAR-10 with a similar number of parameters. In my opinion, the Pyramidal RoR network can not only creates several direct paths for propagating information between different original residual blocks by adding extra shortcuts, so layers in upper blocks can propagate information to layers in lower blocks. By information propagation, it can alleviate the vanishing gradients problem. Moreover, increasing the number of feature map channels gradually ensures the diversity of advanced attributes, while ensuring the continuity of information.

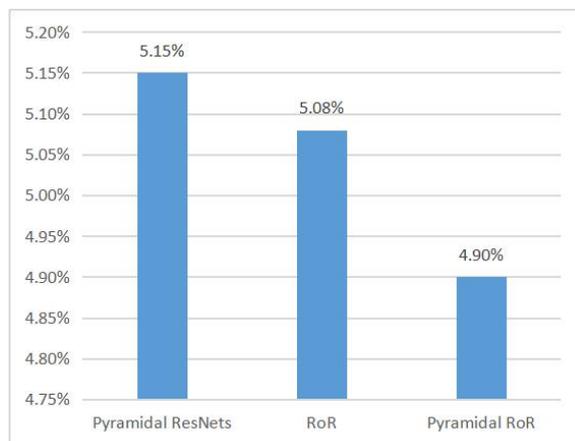

Figure.7 Performance comparison of 110-layer model on CIFAR-10.

The experimental results perfectly validate the effectiveness of the proposed optimization method. In this paper, we think that, though multi-level shortcuts, different grades of features can be connected to each other, further inhibit the gradient disappeared. The number of feature channels increasing can not only increasing the diversity of features, but also making the information more coherent. which means, we obtained better results.

**5.3 Depth and Width Experiments**

Gradient vanish is still the main problem in very deep CNN training. we increase output width of networks to improve the performance, instead of blindly deepened networks (causing the gradient vanishes), which controls the vanish gradient in the same order of magnitude. We repeated the Pyramidal RoR+SD experiments by increasing the number of convolution filters in each layer as shown in Table 3 and Figure 8. The convergence curve is shown in Figure. 8. As can be observed, when increased from 48 to 84, and then to 270 (in 110-layer Pyramidal RoR+SD), the performance increased gradually. Which means Pyramidal RoR model has a good generalization ability to the output dimension. Output dimension increases and TOP-1 error rate decreases. In view of good performance, we tried a deeper model (depth=146, $\alpha$ =270) and got the current lowest classification error rate on CIFAR-10/100. The lowest classification error rates on CIFAR-10/100 were 2.96%, 16.40%, respectively.

Table 3 Test error (%)on Width and Depth experiments

|  | CIFAR-10 | CIFAR-100 |
| --- | --- | --- |
| depth=110, α=48 | 4.35% | 21.41% |
| depth=110, α=84 | 3.99% | 19.58% |
| depth=110, α=270 | 3.33% | 16.82% |
| depth=110, α=270 | 2.96% | 16.40% |

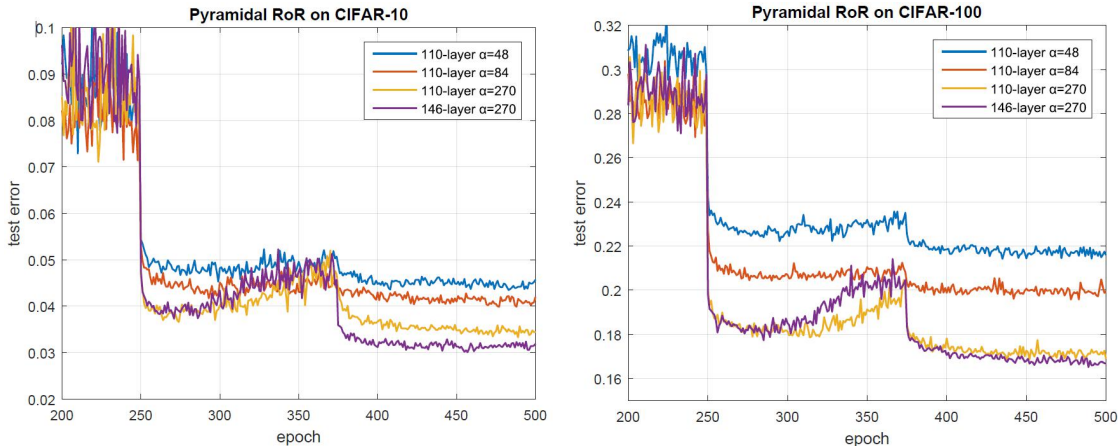

Figure.8 Test Error (%) on 110-layer Pyramidal RoR+SD

### 5.4 Experiments on SVHN

The Street View House Number (SVHN) data set used in this research contains 32 × 32 colored images of cropped out house numbers from Google Street View [20]. The task is to classify the digit at center (and ignore any additional digit that might appear on the side) of the images. There are 73,257 digits in the training set, 26,032 in the test set and 531,131 easier samples for additional training. Following the common practice, we used all the training samples but did not perform data augmentation. We preprocessed the data by subtracting the mean and dividing the standard deviation. In this paper, we chose the excellent model (Pyramidal RoR+SD (depth=110, α=270) and (depth=146, α=270)) to apply on SVHN. Batch size was set to 128, and test error was calculated every 200 iterations. The classification error rates on SVHN are shown in Table 4.

Table 4 Test Error (%)on SVHN experiments

| | SVHN |
|---|---|
| depth=110, α=270 | 1.61 |
| depth=146, α=270 | 1.59 |

### 5.5 Results Comparison of the Best Model Classification

Table 5 compares the state-of-the-art methods on CIFAR-10/100, and we achieved overwhelming results. The reasons for the performance are as follows: Through the multi-level shortcuts, different grades of features can be connected to each other, further inhibit the gradient disappeared. The number of feature channels is increasing gradually, that can not only increase the diversity of features, but also making the information more coherent. Our Pyramid RoR+SD (depth=110, α=48) (1.7M) had an error of 4.35% on CIFAR-10, which was better than the 5.23% of ResNet-110+SD and the 5.08% of RoR-110+SD with the same number of magnitude parameters. Our Pyramidal RoR+SD (depth=110, α=48) (1.7M) had an error of 21.41% on CIFAR-100, which was better than the 24.58% of ResNet-110+SD and the 23.48% of RoR-110+SD with the same number of magnitude parameters. According to the experimental results, it is found that the Pyramidal RoR network achieves a better classification effect in the 110-layer network (1.7M), which shows that the model has better fitting ability in the same parameter. In addition, our model has a good generalization ability for different output dimensions. Particularly, our Pyramid RoR+SD (depth=146, α=270) obtained a single-model error of 2.96% on CIFAR-10, 16.40% on CIFAR-100, which are now state-of-the-art performance standards, to the best of our knowledge. On the SVHN dataset, we also received an overwhelming error rate of 1.59%. Our results are the same as the best results, and no transcending the best results may be due to training initialization with randomness. These results demonstrate the effectiveness and versatility of Pyramidal RoR.

Table 5 Test error (%) comparison of optimal models

| Method (Parameters) | CIFAR-10 | CIFAR-100 | SVHN |
|---|---|---|---|
| Highway [26] | 7.72 | 32.39 | - |
| ELU [24] | 6.55 | 24.28 | - |

| Model | | | |
|---|---|---|---|
| FractalNet(30M) [27] | 4.59 | 22.85 | 1.87 |
| ResNet-164(2.5M) [8] | 5.93 | 25.16 | 1.75 |
| Pre-ResNet-164(2.5M) [17] | 5.46 | 24.33 | - |
| Pre-ResNet-1001(10.2M) [12] | 4.62 | 22.71 | - |
| ELU-ResNets-110 (1.7M) [28] | 5.62 | 26.55 | - |
| PELU-ResNets-110 (1.7M) [25] | 5.37 | 25.04 | - |
| ResNet-110+SD(1.7M) [13] | 5.23 | 24.58 | - |
| ResNet in ResNet (10.3M) [31] | 5.01 | 22.90 | - |
| WResNet-d (19.3M) [15] | 4.70 | - | - |
| WRN28-10 (36.5M) [14] | 4.17 | 20.50 | 1.64 |
| CRMN(>40M) [29] | 4.65 | 20.35 | 1.68 |
| RoR-110+SD (1.7M) [21] | 5.08 | 23.48 | - |
| RoR-WRN56-4（13.3M）[21] | 3.77 | 19.73 | 1.59 |
| multi-resnet（145M）[32] | 3.73 | 19.60 | - |
| DenseNet(27.2M) [22] | 3.74 | 19.25 | 1.59 |
| PyramidNet (28.3M) [34] | 3.77 | 18.29 | - |
| ResNeXt-29, 16×64d (68.1M) [33] | 3.58 | 17.31 | - |
| **Pyramid RoR+SD (depth=110, α=48) (1.7M)** | **4.35** | **21.41** | **-** |
| **Pyramid RoR+SD (depth=110, α=84) (3.8M)** | **3.99** | **19.58** | **-** |
| **Pyramid RoR+SD (depth=110, α=270) (28.3M)** | **3.33** | **16.82** | **1.63** |
| **Pyramid RoR+SD (depth=146, α=270) (38M)** | **2.96** | **16.40** | **1.59** |

On the other hand, while obtaining excellent classification performances, the number of parameters in our network is not much. Although ResNeXt-29, multi-resnet, and CRMN achieve competitive test errors, the number of parameters in these models is too large (as shown in Table 4). Through experiments and analysis, we argue that our Pyramidal RoR can outperform other methods with a similar order of magnitude parameters. Our Pyramidal RoR models with only 1.7M parameters (Pyramidal RoR+SD (depth=110, α=48) (1.7M)) can outperform Pre-ResNet-1001(10.2M), FractalNet (30M), WResNet-d (19.3M) and CRMN-28 (> 40M parameters) on CIFAR-10. Our best Pyramidal RoR+SD (depth=146, α=270) model with 38M parameters achieved the new state-of-the-art performance.

## 6. Conclusion

In this paper, we put forward an optimization method of Residual Networks of Residual Networks (RoR) named Pyramidal RoR network model framework. We acquired state-of-the-art image classification results on CIFAR-10, CIFAR-100 and SVHN. The experiment results show Pyramidal RoR can give more control over bias shift and vanishing gradients and get excellent image classification performance.

**Acknowledgements**

This work is supported by National Natural Science Foundation of China (Grants No. 61302163, No. 61302105 and No. 61501185), Hebei Province Natural Science Foundation (Grants No. F2015502062 and No. F2016502062) and the Fundamental Research Funds for the Central Universities (Grants No. 2016MS99).

**References**
[1] Y. LeCun, Y. Bengio, and G. Hinton, "Deep learning," Nature, vol. 521, no. 7553, pp. 436–444, May. 2015.
[2] W. Y. Zou, X. Y. Wang, M. Sun, and Y. Lin, "Generic object detection with dense neural patterns and regional," *arXiv preprintarXiv:1404.4316*, 2014.
[3] A. Krizhenvshky, I. Sutskever, and G. Hinton, "Imagenet classification with deep convolutional networks," in *Proc. Adv. Neural Inf. Process. Syst.*, 2012, pp. 1097–1105.
[4] O. Russakovsky, J. Deng, H. Su, J. Krause, S. Satheesh, S. Ma, Z. Huang, A. Karpathy, A. Khosla, M. Bernstein, A. C. Berg, and L. Fei-Fei, "Imagenet large scale visual recognition challenge," *arXiv preprint arXiv:1409.0575*, 2014.
[5] Y. Bengio, P. Simard, and P. Frasconi, "Learning long-term dependencies with gradient descent is difficult," *IEEE Trans. Neural Networks*, vol. 5, no. 2, pp. 157–166, Aug. 2014.
[6] K. Simonyan, and A. Zisserman, "Very deep convolutional networks for large-scale image recognition," *arXiv preprint arXiv:1409.1556*, 2014.
[7] C. Szegedy, W. Liu, Y. Jia, P. Sermanet, S. Reed, D. Anguelov, D. Erhan, V. Vanhoucke, and A. Rabinovich, "Going deeper with convolutions," in *Proc. IEEE Conf. Comput. Vis. Pattern Recog.*, 2015, pp. 1–9.
[8] K. He, X. Zhang, S. Ren, and J. Sun, "Deep residual learning for image recognition," *arXiv preprint arXiv:1512.03385*, 2015.
[9] S. Ioffe, and C. Szegedy, "Batch normalization: accelerating deep network training by reducing internal covariate shift," *arXiv preprint arXiv:1502.03167*, 2015.
[10] D. Mishkin, and J. Matas, "All you need is a good init," *arXiv preprint arXiv:1511.06422*, 2015.

[11] K. He, and J. Sun, "Convolutional neural networks at constrained time cost," in *Proc. IEEE Conf. Comput. Vis. Pattern Recog.*, 2015, pp. 5353–5360.
[12] K. He, X. Zhang, S. Ren, and J. Sun, "Identity mapping in deep residual networks," *arXiv preprint arXiv:1603.05027*, 2016.
[13] G. Huang, Y. Sun, Z. Liu, and K. Weinberger, "Deep networks with stochastic depth," *arXiv preprint arXiv:1605.09382*, 2016.
[14] S. Zagoruyko, and N. Komodakis, "Wide residual networks," *arXiv preprint arXiv:1605.07146*, 2016.
[15] F. Shen, and G. Zeng, "Weighted residuals for very deep networks," *arXiv preprint arXiv:1605.08831*, 2016.
[16] G. Hinton, N. Srivastava, A. Krizhevsky, and K. Weinberger, "Improving neural networks by preventing co-adaptation of feature detectors," *arXiv preprint arXiv:1207.0580*, 2012.
[17] N. Srivastava, G. Hinton, A. Krizhevsky, I. Sutskever, and R. Salakhutdinov, "Dropout: a simple way to prevent neural networks from overfitting," *The Journal of Machine Learning Research*, vol. 15, pp. 1929–1958, Jun. 2014.
[18] S. Ioffe, and C. Szegedy, "Batch normalization: accelerating deep network training by reducing internal covariate shift," *arXiv preprintarXiv:1502.03167*, 2015.
[19] A. Krizhenvshky, and G. Hinton, "Learning multiple layers of features from tiny images," M.Sc. thesis, Dept. of Comput. Sci., Univ. of Toronto, Toronto, ON, Canada, 2009.
[20] Y. Netzer, T. Wang, A. Coates, A. Bissacco, B. Wu, and A. Y. Ng, "Reading digits in natural images with unsupervised feature learning," in *Proc. NIPS Workshop Deep Learning and Unsupervised feature learning.*, 2011, pp. 1–9.
[21] Zhang K, Sun M, Han X, et al. Residual Networks of Residual Networks: Multilevel Residual Networks[J]. IEEE Transactions on Circuits Systems for Video Technology, 2016, PP (99):1-1.
[22] Huang G, Liu Z, Weinberger K Q, et al. "Densely connected convolutional networks," *arXiv preprint arXiv:1608.06993*, 2016.
[23] V. Nair, and G. Hinton, "Rectified linear units improve restricted Boltzmann machines," in *Proc. ICML*, 2010, pp. 807–814.
[24] D. -A. Clevert, T. Unterthiner, and S. Hochreiter, "Fast and accurate deep network learning by exponential linear units (elus)," *arXiv preprint arXiv:1511.07289*, 2015.
[25] L. Trottier, P. Giguere, and B. Chaib-draa, "Parametric exponential linear unit for deep convolutional neural networks," *arXiv preprint arXiv:1605.09322*, 2016.
[26] R. K. Srivastava, K. Greff, and J. Schmidhuber, "Highway networks," *arXiv preprint arXiv:1505.00387*, 2015.
[27] G. Larsson, M. Maire, and G. Shakhnarovich, "FractalNet: ultra-deep neural networks without residuals," *arXiv preprint arXiv:1605.07648*, 2016.
[28] A. Shah, S. Shinde, E. Kadam, and H. Shah, "Deep residual networks with exponential linear unit," *arXiv preprint arXiv:1604.04112*, 2016.
[29] J. Moniz, and C. Pal, "Convolutional residual memory networks," *arXiv preprint arXiv:1606.05262*, 2016.
[30] G. Hinton, N. Srivastava, A. Krizhevsky, and K. Weinberger, "Improving neural networks by preventing co-adaptation of feature detectors," *arXiv preprint arXiv:1207.0580*, 2012.
[31] S. Targ, D. Almeida, and K. Lyman, "Resnet in resnet: generalizing residual architectures," *arXiv preprint arXiv:1603.08029*, 2016.
[32] Abdi M, Nahavandi S, "Multi-residual networks," *arXiv preprint arXiv:1609.05672*, 2016.
[33] Xie S, Girshick R, Dollr P, et al, "Aggregated residual transformations for deep neural networks," *arXiv preprint arXiv:1611.05431*, 2016.
[34] Han D, Kim J, Kim J, "Deep pyramidal residual networks," in *Proc. CVPR.*, 2017.
[35] Chen Y, Li J, Xiao H, et al, "Dual Path Networks," in *Proc. CVPR.*, 2017.
[36] Wang F, Jiang M, Qian C, et al, "Residual Attention Network for Image Classification," in *Proc. CVPR.*, 2017.